\PassOptionsToPackage{table}{xcolor}
\documentclass[runningheads]{llncs}

\usepackage[final,year=2026]{eccv}
\usepackage{eccvabbrv}
\usepackage{graphicx}
\usepackage{booktabs}
\usepackage{caption}
\usepackage{amsmath}
\usepackage{amssymb}
\usepackage{array}
\usepackage{xcolor}
\usepackage{colortbl}
\definecolor{OursGray}{gray}{0.88}
\definecolor{DelayGray}{gray}{0.94}
\definecolor{IssueRed}{RGB}{255,225,225}
\usepackage[accsupp]{axessibility}
\usepackage{hyperref}
\usepackage{enumitem}

\usepackage{xcolor}

\definecolor{paulcolor}{RGB}{200, 30, 30}    
\definecolor{ekkasitcolor}{RGB}{30, 100, 200}   

\begin{document}

\title{InteractGesture: Progressive Chunk Guidance for Continuous Streaming Co-Speech Gesture Control}
\titlerunning{InteractGesture}

\author{Ekkasit Pinyoanuntapong\inst{1,2} \and Ajinkya Deogade\inst{1} \and Paul Streli\inst{1} \and Wenjing Zhang\inst{1} \and Joanna Materzynska\inst{1} \and Pu Wang\inst{2} \and Vittorio Ferrari\inst{1} \and Jie Shen\inst{1}}
\authorrunning{E. Pinyoanuntapong et al.}
\institute{Meta \and University of North Carolina at Charlotte}

\maketitle

\begin{figure*}[ht]
\vspace{-25pt}
\centering
\includegraphics[width=\textwidth]{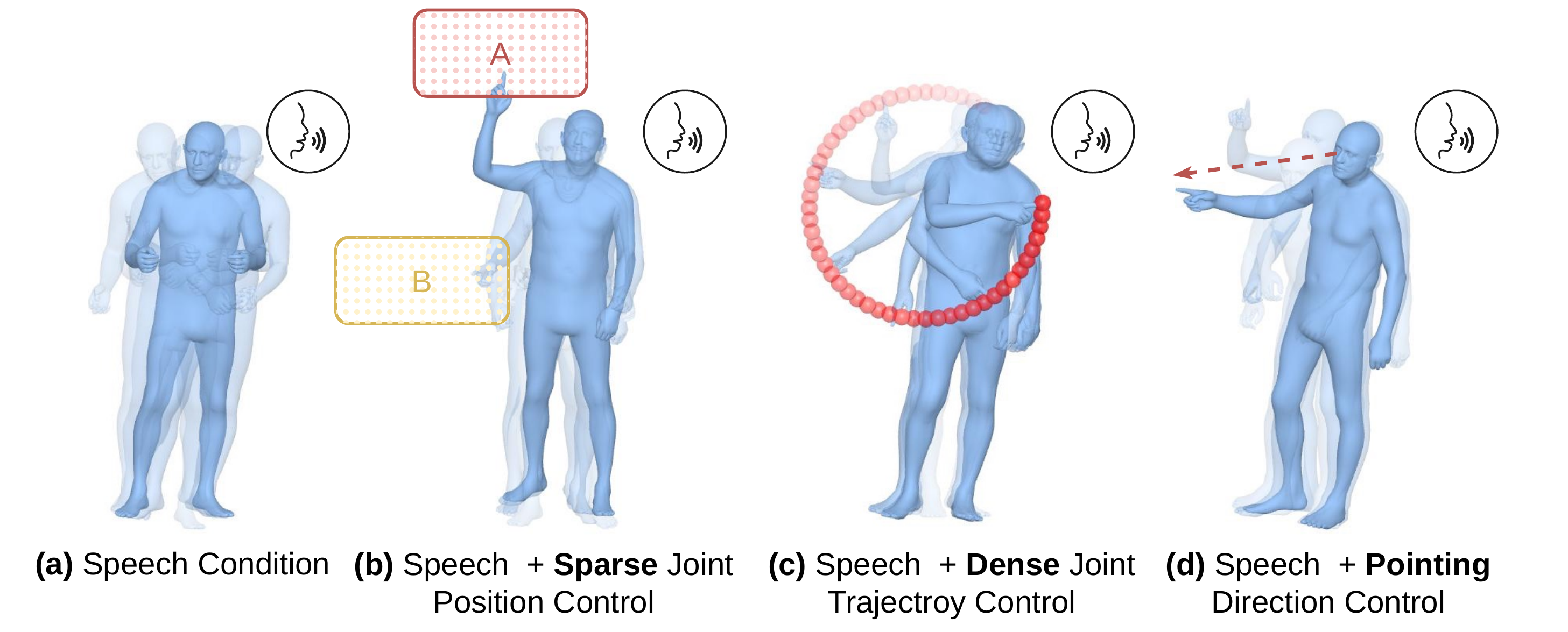}
\vspace{-20pt}
\captionsetup{font=footnotesize}
\caption{\scriptsize \emph{InteractGesture} enables spatial gesture control for a speech-conditioned gesture generation model. (a) Speech Condition Only. (b) Joint Position Control: directs specified joints to point at described objects at designated frames. (c) Dense Joint Trajectory Control: guides joints along a continuous path for sophisticated gesture control. (d) Pointing Direction Control: orients the arm toward a target direction.}
\label{fig:landing}
\vspace{-30pt}
\end{figure*}

\begin{abstract}
Co-speech gesture generation has made significant progress toward realistic full-body motion from speaker audio, yet existing models lack fine-grained spatial controllability of individual joints.
To address this, we introduce \emph{InteractGesture}, a model-agnostic, inference-time method for spatially controllable gesture generation.
\emph{InteractGesture} guides target latent estimates of a diffusion sampler through a differentiable RVQ-VAE decoder, backpropagating spatial control gradients to adjust motion latents during sampling.
A primary challenge in streaming co-speech generation is chunk-wise dependency: standard sequential inference freezes prior chunks, preventing spatial constraints in future chunks from adjusting preceding trajectories and causing boundary inconsistencies.
To overcome this limitation, we propose \emph{Progressive Chunk Guidance}, a chunk-window strategy that maintains an active set of editable chunk latents with staggered delays, enabling spatial constraints to propagate gradients backward across chunk boundaries during streaming generation.
Experiments on the BEAT2 dataset show that \emph{InteractGesture} improves multi-joint spatial control while preserving overall gesture quality.
Furthermore, our approach supports diverse applications, including sparse joint positioning, dense joint trajectory control, and directional pointing.
Our project page is available at \url{https://exitudio.github.io/interactgesture-page}.
\end{abstract}


\section{Introduction}

Speech-driven gesture synthesis is an essential capability for digital humans, virtual agents, and embodied assistants.
The task is challenging because movements must be synchronized with speech rhythm, semantically compatible with spoken content, and natural and coherent across the upper body, hands, lower body, and global root motion.
Previous approaches have improved realism through larger datasets, increasingly powerful temporal models, expressive body representations, and generative gesture priors~\cite{BEAT,EMAGE,DiffGesture,GestureLSM}.
Yet these methods are still primarily designed for unconstrained generation: given speech and optional style information, they produce plausible gestures, but do not provide fine-grained spatial controllability of individual body parts.

Spatial controllability is crucial for practical gesture editing and embodied interaction.
An animator may want the right wrist to land near a prop, both hands to frame an object during an explanation, or an elbow trajectory to avoid an obstacle.
An embodied agent may need to point toward an interface element while continuing to gesture naturally with speech.
Effective spatial control should preserve speech-gesture synchrony while enabling precise direct control over generated motions (see Figure~\ref{fig:landing}).

We introduce \emph{InteractGesture}, a model-agnostic, inference-time method for spatially controllable co-speech gesture generation.
\emph{InteractGesture} treats a pretrained gesture generator as a denoising process whose target latent estimates can be guided during inference.
At selected sampling steps, the predicted target latent estimate is decoded through a differentiable RVQ-VAE decoder into body poses, a masked spatial loss is computed against user-specified control points, and gradients update the target latent estimate rather than the model parameters.
The adjusted estimate is then used to denoise the current motion latent, allowing subsequent sampling steps to pull the guided state back toward the learned speech-conditioned gesture distribution.

Prior work in controllable motion synthesis~\cite{GMD, OmniControl, ProgrammableMotionGeneration, MaskControl} demonstrates that generative models can follow spatial trajectories, keyframes, or programmable constraints.
However, these methods operate on text-to-motion tasks, where the conditioning signal is static and the output is typically a short, offline clip.
Co-speech motion synthesis requires a fundamentally different inference paradigm: gestures must accompany long speech sequences whose continuous audio is typically processed in finite, overlapping chunks, requiring spatial controls to remain consistent across chunk boundaries.

\begin{figure*}[t]
\centering
\includegraphics[width=1\textwidth]{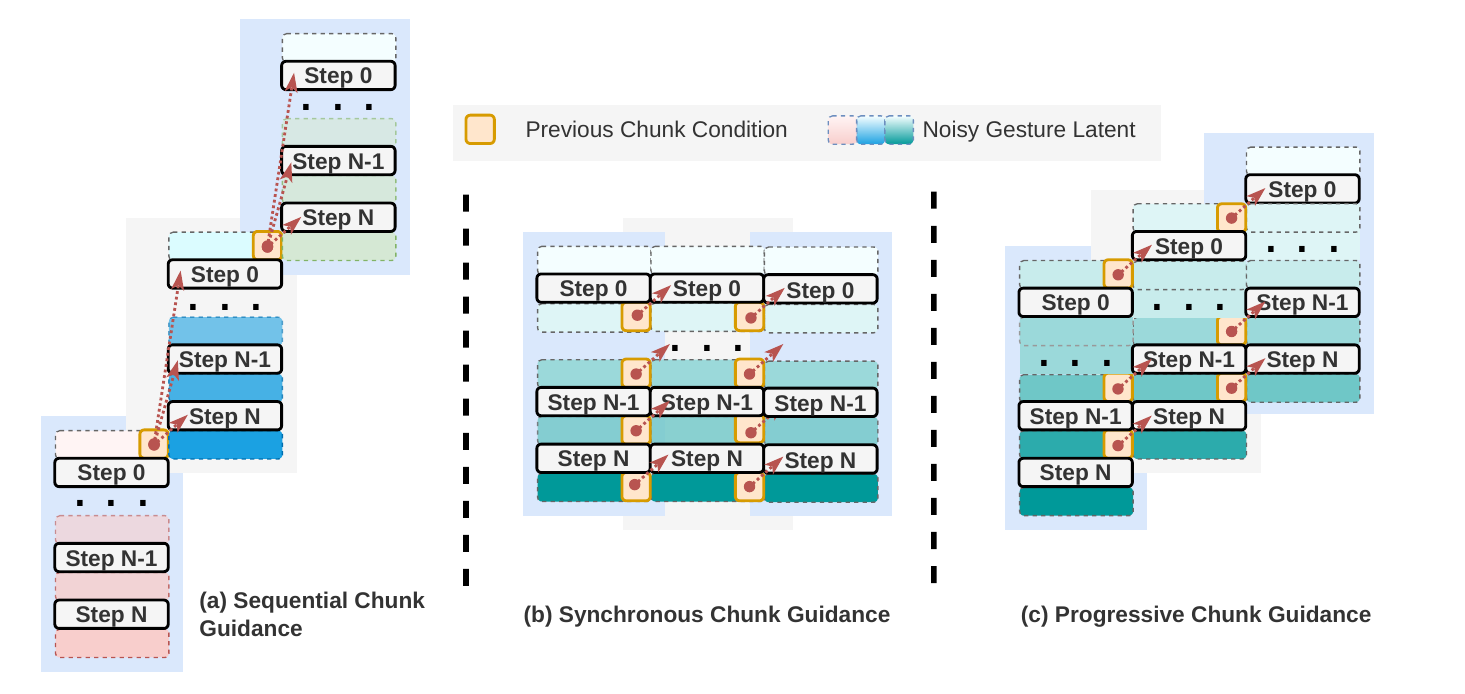}
\caption{
Comparison of chunk-guidance schedules.
\emph{(a) Sequential Chunk Guidance:} Fully completes and freezes chunk $k$ before initializing chunk $k+1$. While causal, guidance from later chunks cannot adjust earlier motion.
\emph{(b) Synchronous Chunk Guidance:} Denoises all chunks simultaneously. Controls across the entire sequence influence overlapping context, but this requires full-sequence audio offline.
\emph{(c)} \emph{Progressive Chunk Guidance} (Ours): Introduces chunks with a staggered delay and continuously refreshes context frames from intermediate target estimates $\hat{\mathbf{x}}_0$. This maintains streaming compatibility while propagating spatial control across chunk boundaries.
}
\label{fig:chunk_compare}
\end{figure*}

For standard chunk-wise inference, chunk $k$ must be fully generated before chunk $k+1$ can use it as historical context.
This strict \emph{sequential} dependency complicates spatial control: reaching a target in chunk $k+1$ may require chunk $k$ to adjust its trajectory, but chunk $k$ is already locked (see Figure~\ref{fig:chunk_compare}a).
Therefore, independently guiding chunks produces inconsistent control effects across boundaries.

To avoid this, given the full audio stream, all chunks can be denoised \emph{synchronously} (see Figure~\ref{fig:chunk_compare}b).
This allows spatial controls to propagate gradients backward across earlier chunk latents during intermediate sampling steps, adjusting preceding trajectories before they are finalized.

To combine the online capability of chunk-wise generation with the global coherence of joint optimization, we propose \emph{Progressive Chunk Guidance}.
Rather than fixing past context or waiting for entire sequences to finalize, \emph{InteractGesture} maintains an active window of editable chunk latents that are denoised concurrently.
As continuous audio streams in, newly active chunks are introduced with staggered delays and conditioned on the evolving, guided latent estimates of preceding chunks (see Figure~\ref{fig:chunk_compare}c).
This continuous backward propagation of gradients across active chunks enables future spatial controls to adjust preceding motion, reconciling long-horizon controllability with online streaming execution.

Our method is model-agnostic at the control interface: it does not depend on specific training objective or generator architecture, provided the model exposes an editable target latent estimate (or equivalent latent update) and a differentiable decoder mapping latents to joint poses.
Our implementation uses the pretrained RVQ-VAE body-part decoders and SMPL-X joint recovery from a co-speech generator, which provides an end-to-end differentiable gradient path from a joint-space control loss back to the latent estimates.
The same representation supports diverse control tasks, including absolute scene-space targets, root-relative body-centric targets, sparse joint positioning, dense trajectory control, and directional pointing.\\

\noindent Collectively, we contribute:
\begin{enumerate}[topsep=4pt, itemsep=2pt, parsep=0pt]
    \item \emph{InteractGesture}, a model-agnostic, inference-time framework for fine-grained spatial control of pretrained co-speech gesture generators.
    \item \emph{Progressive Chunk Guidance}, a chunk-window sampling strategy that maintains editable active chunks to enforce spatial control consistency across streaming chunk boundaries.
    \item An evaluation on the BEAT2 dataset with quantitative and qualitative results, demonstrating the effectiveness of \emph{InteractGesture} and its diverse applications, including sparse joint positioning, dense trajectory tracking, and directional pointing.
\end{enumerate}

\section{Related Work}

\subsection{Co-Speech Gesture Generation}

Co-speech gesture generation synthesizes body motion from speech audio, text, speaker identity, and conversational context.
Early work learned multimodal mappings for upper-body motion or video-based gesture transfer~\cite{ginosar2019gestures,TrimodalGesture,SpeechGestureGeneration,EverybodyDanceNow,wang2018pix2pixHD}, while later skeleton-based methods improve speech-gesture alignment with stronger audio-motion representations, semantic cues, contrastive objectives, and conversational context~\cite{liu2022learning,ao2022rhythmic,xu2023chaingenerationmultimodalgesture,liu2024tango,zhang2024kinmokinematicawarehumanmotion,liu2025contextualgesturecospeechgesture,liu2025intentionalgesturedeliverintentions,liu2025semgessemanticsawarecospeechgesture,semanticgesticulator}.
Dataset and representation advances such as BEAT/CaMN~\cite{BEAT,liu2022beat}, BEAT2~\cite{EMAGE}, TalkSHOW~\cite{yi2022generating,SMPLX}, EMAGE~\cite{EMAGE}, and latent representations such as VQ-VAE/RVQ-VAE~\cite{VQVAE,RVQVAE,GestureLSM} further broaden the task from upper-body prediction to holistic body, hand, and facial generation.

Recent work also explores diffusion and masked generative modeling~\cite{DiffGesture,Deichler_2023,diffsheg,chen2024syntalker}, personalization and emotion control~\cite{PersonaGesture,PersonaGest,EmotionGesture,EmoDiffGes,AMUSE}, semantic planning~\cite{GesGPT}, photoreal rendering~\cite{Audio2Photoreal}, and dyadic or spatially aware interaction~\cite{DyaDiT,MIBURI,SARAH,DualTalk,DyaPlex}.
These methods broaden speech-driven generation, but their controls are primarily identity, style, emotion, semantic intent, or interaction context rather than direct 3D spatial targets for individual joints.
For streaming generation, CaMN-style autoregressive models~\cite{liu2022beat}, MambaTalk~\cite{mambatalk}, and LiveGesture~\cite{LiveGesture} generate gestures from incoming audio.
\emph{InteractGesture} shares this streaming motivation, but focuses on fine-grained inference-time spatial control while chunks are still being generated.

\subsection{Controllable Human Motion Generation}

Text-to-motion control methods add trajectories, inbetweening, body-part masks, programmable constraints, or joint keyframes to offline motion generation~\cite{GMD,OmniControl,ProgrammableMotionGeneration,MaskControl,MMM,BAMM,MoMask}.
These methods show that generative motion priors can follow spatial constraints, but they typically assume static text prompts, fixed sequence lengths, and offline generation.
Controllable co-speech work has begun to study pointing, trajectory control, multimodal examples, and object-aware interaction~\cite{SpatialAwarenessGesture,TrajectoryControllableGesture,MECo,InteracTalker}, yet it does not provide fine-grained multi-joint 3D spatial control during continuous streaming generation.

At the algorithmic level, diffusion and flow samplers can be steered with classifier guidance, classifier-free guidance, energy guidance, or optimization-based editing~\cite{DDPM,ClassifierFreeGuidance,lipman2022flow}.
For human motion, differentiable kinematics and body models allow geometric losses to be evaluated on decoded poses during sampling.
\emph{InteractGesture} adapts this principle to co-speech latent generation by optimizing target latent estimates through a differentiable decoder, while \emph{Progressive Chunk Guidance} propagates these edits across active streaming chunk boundaries.

\section{Method}
\label{sec:method}

\begin{figure*}[t]
\centering
\includegraphics[width=1\textwidth]{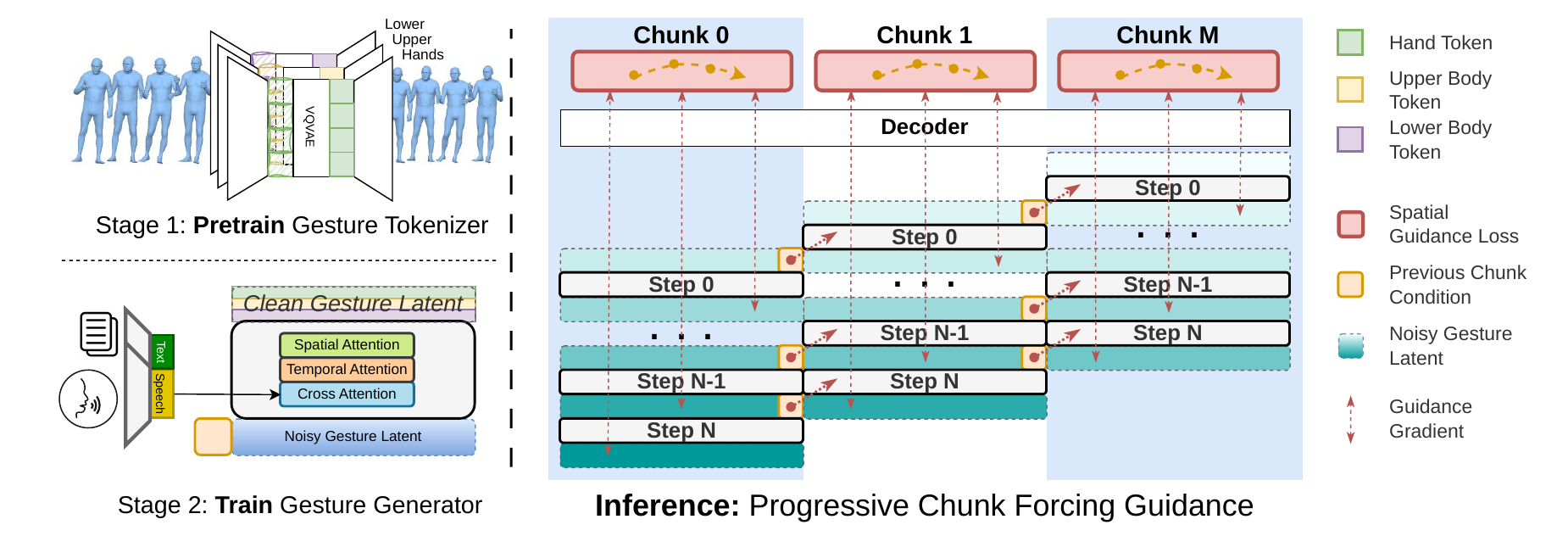}
\caption{
\emph{InteractGesture} adds spatial control guidance to a pretrained co-speech gesture generation model at inference time.
At every selected sampling step, the predicted target latent estimate is decoded into SMPL-X joints, a masked joint loss is minimized, and the edited target latent estimate is converted back into the sampler update used by the generator.
The same mechanism supports absolute 3D targets, root-relative targets, and \emph{Progressive Chunk Guidance} for continuous streaming generation from overlapping latent chunks.
}
\label{fig:architecture}
\end{figure*}


Our method synthesizes speech-synchronized full-body gestures while enforcing spatial constraints during continuous streaming generation.
\emph{InteractGesture} augments a pretrained chunk-level latent sampler conditioned on audio with a spatial control specification $\mathcal{C}=(\mathbf{H},\mathbf{M},\Phi)$, where $\mathbf{H}$ stores target joint positions, $\mathbf{M}$ denotes active joint-frame masks, and $\Phi$ projects generated skeleton joints into the target coordinate space.
Treating the underlying generator as a black box, our formulation backpropagates control gradients exclusively into the target latent estimates $\hat{\mathbf{z}}$, leaving model parameters untouched.

\subsection{Pretrained Latent Generator}

Our pretrained motion generator synthesizes streaming motion as a sequence of motion chunks using a fixed-step DDIM sampler. Each generated motion chunk contains 128 pose frames. Because the underlying RVQ-VAE uses a temporal squeeze factor of 4, the sampler operates on latent windows of length 32.
The latent channel is partitioned into three body-part streams:
\begin{equation}
    \mathbf{z} = \left[ \mathbf{z}^{\mathrm{up}},\, \mathbf{z}^{\mathrm{hand}},\, \mathbf{z}^{\mathrm{low}} \right] \in \mathbb{R}^{32 \times 384},
\end{equation}
where each stream $\mathbf{z}^{m} \in \mathbb{R}^{32 \times 128}$ corresponds to upper-body, hand, or lower-body dynamics, respectively.
The upper-body decoder reconstructs upper-body 6D rotations, the hand decoder reconstructs hand joint rotations, and the lower-body decoder predicts lower-body rotations alongside root translation velocity.
Global root translation $\mathbf{r}$ is recovered by cumulative summation of the predicted velocities across temporal frames.

These decoded body-part poses are mapped into their corresponding joint indices within the full $55$-joint SMPL-X pose vector.
Given body shape parameters $\boldsymbol{\beta}$ and global root translation $\mathbf{r}$, the end-to-end differentiable SMPL-X layer computes 3D joint positions:
\begin{equation}
    \mathbf{J}(\mathbf{z}; \boldsymbol{\beta}) = \operatorname{SMPLX}\left( \operatorname{Dec}_{\mathrm{up}}(\mathbf{z}^{\mathrm{up}}),\, \operatorname{Dec}_{\mathrm{hand}}(\mathbf{z}^{\mathrm{hand}}),\, \operatorname{Dec}_{\mathrm{low}}(\mathbf{z}^{\mathrm{low}}),\, \mathbf{r},\, \boldsymbol{\beta} \right).
\end{equation}
This fully differentiable decoding path enables spatial control loss gradients to backpropagate directly from 3D joint space into the target latent estimates $\mathbf{z}$, serving as the primary interface required for guidance.

\subsection{Control Representation}

\emph{InteractGesture} specifies spatial controls using a target position tensor and a corresponding binary indicator mask.
For a gesture sequence of length $T$ frames and $J = 55$ SMPL-X joints, the spatial hint tensor is defined as:
\begin{equation}
    \mathbf{H} \in \mathbb{R}^{T \times J \times 3},
\end{equation}
and the spatial mask tensor as:
\begin{equation}
    \mathbf{M} \in \{0,1\}^{T \times J},
\end{equation}
where entries $\mathbf{M}_{t,j} = 1$ denote active spatial constraints contributing to the control loss, while $\mathbf{M}_{t,j} = 0$ indicates unconstrained joint-frame positions.
This tensorized formulation naturally accommodates arbitrary spatio-temporal constraint configurations, ranging from sparse joint keyframes to continuous multi-joint trajectories.


To enforce constraints during continuous streaming generation with overlapping windows, let $P$ denote the pose chunk length ($P=128$ frames) and $S$ denote the number of historical context frames.
The non-overlapping stride is given by $R = P - S$.
For the $k$-th chunk, spatial constraints from the sequence-level hint tensor are sliced across local frame indices $kR$ to $kR + P$.
For all non-initial chunks ($k > 0$), constraints that overlap with the historical context region ($1 \le t \le S$) are masked out in $\mathbf{M}$.
This prevents double-counting optimization loss on previously generated context while preserving a unified, sequence-level control specification across streaming chunk boundaries.





\subsection{Guidance on Target Latent Estimates}

The DDIM sampler maintains intermediate latent states along a reverse denoising trajectory $\mathbf{x}_N, \dots, \mathbf{x}_0$ over denoising steps $i \in \{N, \dots, 1\}$.
At denoising step $i$, the generator predicts the clean target latent estimate $\hat{\mathbf{x}}_0$.
Our spatial control loss evaluates this prediction across motion frames $t$ and joints $j$:
\begin{equation}
    \mathcal{L}_{\mathrm{ctrl}}(\hat{\mathbf{x}}_0)
    =
    \frac{1}{2}
    \sum_{b,t,j}
    \mathbf{M}_{b,t,j}
    \left\|
    \Phi\left( \mathbf{J}(\hat{\mathbf{x}}_{0,b})_{t,j} \right) - \mathbf{H}_{b,t,j}
    \right\|_2^2.
\end{equation}

The optimization variable is $\hat{\mathbf{x}}_0$ itself.
All generator parameters, RVQ-VAE decoders, and SMPL-X body model weights remain fixed.

We optimize $\hat{\mathbf{x}}_0$ using the Adam optimizer.
Because the number of active constraints varies across joint groups and keyframe densities, we normalize the base learning rate $\eta_0$ by the active constraint count $N_{\mathrm{active}}$:
\begin{equation}
    \eta = \frac{\eta_0}{g(N_{\mathrm{active}})},
\end{equation}
where $g(N_{\mathrm{active}}) = N_{\mathrm{active}}$ for linear normalization, $g(N_{\mathrm{active}}) = \sqrt{N_{\mathrm{active}}}$ for square-root normalization, and $g(N_{\mathrm{active}}) = 1$ when unnormalized.
By default, we use linear normalization.

Following gradient optimization, the edited target estimate $\hat{\mathbf{x}}_0^{*}$ is plugged directly into the standard denoising step formula to update the next latent state $\mathbf{x}_{i-1}$.
While spatial edits temporarily push the estimated trajectory away from the unconstrained motion distribution, subsequent denoising steps sample from this guided state, allowing the learned score network to continually regularize the motion back toward the pretrained gesture prior.
Because this framework relies solely on optimizing the estimated target state $\hat{\mathbf{x}}_0$, this formulation naturally extends to other generative paradigms, such as Flow Matching and standard DDPM samplers, without architectural modification.



\subsection{Guidance Schedule}

Not all reverse sampling steps require equal optimization effort.
In early sampling steps, decoded intermediate joints are noisy, making spatial gradients unstable.
In later sampling steps, the latent estimate converges toward the clean target manifold, enabling more precise spatial edits.
To reflect this dynamic, we schedule the number of latent optimization iterations $K(i)$ over the reverse sampler index $i \in \{N, \dots, 1\}$ as:
\begin{equation}
    K(i) = \left\lfloor K_{\mathrm{early}} + (K_{\mathrm{late}} - K_{\mathrm{early}}) \cdot \max\left(0, \min\left(1, \frac{\tau - i}{\tau}\right)\right) \right\rceil,
    \label{eq:schedule}
\end{equation}
where $\tau$ denotes the late-start threshold in denoising step units, $K_{\mathrm{early}}$ and $K_{\mathrm{late}}$ specify the initial and final iteration budgets, and $\lfloor \cdot \rceil$ denotes rounding to the nearest integer.
This schedule concentrates spatial guidance toward the final stage of the sampling trajectory, where joint decoding is most reliable.

Additionally, \emph{InteractGesture} supports an optional post-sampling refinement stage.
After completing the final denoising step, the spatial control loss $\mathcal{L}_{\mathrm{ctrl}}$ can be optimized directly on the final clean target latent estimate $\mathbf{x}_0$ for $K_{\mathrm{post}}$ iterations.
While this refinement further reduces spatial position error, excessive post-sampling optimization bypasses the generative sampler's regularization steps and can diminish gesture naturalness.


\subsection{Progressive Chunk Guidance}
\label{sec:progressive_chunk_guidance}

Continuous gesture generation processes audio incrementally via sliding temporal windows of stride $R = P - S$ frames.
In standard unconstrained rollouts, chunk $k$ is fully generated before its $S$ historical context frames are copied to initialize chunk $k+1$.
While suitable for real-time streaming, this rigid boundary poses a challenge for spatial control: a constraint in chunk $k+1$ may require an earlier arm trajectory adjustment in chunk $k$, which is already frozen.
Consequently, independently guided chunks can satisfy local targets while inducing motion discontinuities across chunk boundaries.


To address this, we compare three chunk-guidance schedules illustrated in Figure~\ref{fig:chunk_compare}: a baseline Sequential strategy, an offline Synchronous scheme, and our proposed \emph{Progressive Chunk Guidance}.


\paragraph{Sequential chunk guidance.}
This causal baseline completes and freezes chunk $k$ before using its final clean context frames to initialize chunk $k+1$.
Because chunk $k$ is immutable when chunk $k+1$ begins denoising, spatial guidance in chunk $k+1$ cannot adjust the incoming motion, leading to abrupt trajectory changes across window boundaries.

\paragraph{Synchronous Chunk Guidance.}
When the entire audio stream is available offline, all chunks are activated from the start and guided concurrently.
Spatial loss gradients backpropagate through all active chunks simultaneously, allowing future constraints to adjust preceding trajectories across overlapping context regions.
While this provides the highest spatial accuracy and trajectory smoothness, it cannot support streaming generation.


\paragraph{Progressive Chunk Guidance.}
For continuous streaming generation, future chunks arrive incrementally.
\emph{Progressive Chunk Guidance} introduces subsequent chunks with a staggered offset of $\Delta i$ sampler steps.
While chunk $k+1$ waits to initialize, preceding chunk $k$ continues to optimize its target latent estimate $\hat{\mathbf{x}}_0^{(k)}$ under spatial guidance.
Crucially, at each active denoising step $i$, chunk $k+1$ refreshes its $S$ context frames from the latest guided estimate of chunk $k$.
This creates a two-way control cascade: spatial loss gradients in chunk $k+1$ flow backward into $\hat{\mathbf{x}}_0^{(k)}$ across the decoder's temporal receptive field, while updated estimates in chunk $k$ dynamically refresh the context consumed by chunk $k+1$.
This bidirectional propagation allows future spatial constraints to adjust preceding arm trajectories, enforcing smooth inter-chunk motion continuity within a fixed, user-defined streaming latency budget.



\subsection{Applications}

The spatial control tensor $\mathbf{H}$ supports three key editing modes:
\emph{Joint Position Control}, Figure~\ref{fig:landing}(b), activates a small number of joint-frame targets, such as placing a wrist at a scene position during a spoken reference.
This serves as our primary quantitative benchmark.
\emph{Dense Trajectory Control}, Figure~\ref{fig:landing}(c), activates targets across continuous frame sequences to enforce hand or elbow paths using a temporally dense mask $\mathbf{M}$.
\emph{Pointing Direction Control}, Figure~\ref{fig:landing}(d), maps pointing rays or arm vector angles to target coordinates along a directional vector, enabling expressive pointing while maintaining speech rhythm.

\section{Experiments}
\label{sec:experiments}

\subsection{Dataset and Evaluation Setup}
\paragraph{Base Generator \& Feature Space.} We conduct evaluation experiments on the BEAT2 dataset using GestureLSM as the pretrained co-speech motion generator.
All GestureLSM generator weights and RVQ-VAE body-part decoders remain frozen throughout evaluation.
Motion is synthesized at 30 FPS using 128-frame pose chunks with a fixed-step DDIM reverse sampling schedule.

\paragraph{Control Benchmarks \& Settings.}
We evaluate spatial control using absolute 3D joint targets extracted from ground-truth SMPL-X poses.
Evaluation covers three constraint densities ($1$, $2$, and $5$ keyframe targets per chunk) across five joint configurations: left wrist, right wrist, left elbow, right elbow, and all four joints combined, yielding 15 distinct control settings.
Guidance optimization executes $K_{\mathrm{late}} = 30$ iterations during late-stage reverse sampling steps, followed by $K_{\mathrm{post}} = 40$ post-sampling refinement iterations on target latents $\hat{\mathbf{x}}_0$.

\paragraph{Evaluated Schedules.}
We compare three chunk-guidance schedules:
(i)~\emph{Sequential Chunk Guidance} (causal baseline freezing prior chunks),
(ii)~\emph{Synchronous Chunk Guidance} (offline upper bound guiding all chunks concurrently), and
(iii)~\emph{Progressive Chunk Guidance} (our proposed streaming schedule using a staggered offset of $\Delta i = 1$ sampling wave).


\subsection{Control Protocol}


The evaluation benchmark is constructed directly from ground-truth SMPL-X poses.
For each sequence, target global translations and joint rotations are converted to 3D joint positions, which are then sampled according to the selected joint group and keyframe density.
All reported metrics are aggregated over all controlled joints and active constraint frames.

\subsection{Metrics}

We evaluate motion quality and spatial control accuracy:
\begin{itemize}[leftmargin=*]
    \item \emph{Fréchet Gesture Distance (FGD $\downarrow$)} measures the distributional feature distance to ground-truth motion.
    \item \emph{Beat Consistency (BC $\rightarrow$)} evaluates the temporal synchronization between speech beats and motion beats (closer to Ground-Truth is preferred).
    \item \emph{Diversity ($\rightarrow$)} evaluates the global pose variation across generated sequences (closer to Ground-Truth is preferred).
    \item \emph{Average Control Error ($\text{Avg. Err. cm} \downarrow$)} is the mean Euclidean distance (in centimeters) between controlled joints and spatial targets.
    \item \emph{Location Error ($\text{Loc. Error} \downarrow$)} is the fraction of keyframe targets with Euclidean error exceeding $10\,\text{cm}$.
    \item \emph{Trajectory Error ($\text{Traj. Error} \downarrow$)} is the fraction of active chunks containing at least one target exceeding $10\,\text{cm}$ error.
\end{itemize}


\subsection{Implementation Details}

The guidance function receives latent tensors from the sampler, re-formats them into temporal chunk windows, and refreshes overlapping context from the latest guided estimates when needed.
Active latents are decoded into body-part streams to extract joint coordinates for loss evaluation. For root-relative control, root rotations are applied prior to loss calculation.

Optimization is performed on latent tensors using Adam.
We configure $K_{\mathrm{early}}=0$, $K_{\mathrm{late}}=30$, and start threshold $\tau=100$, followed by $K_{\mathrm{post}}=40$ post-sampling refinement steps.
We use linear step-size normalization ($\eta = \eta_0 / N_{\mathrm{active}}$) to normalize gradient magnitudes between single-joint controls and dense multi-joint groups.
Without this normalization, dense multi-joint controls produce excessively large gradient updates that distort gesture naturalness.

\subsection{Main Quantitative Results}

\begin{table*}[t]
\centering
\caption{Main spatial control comparison on BEAT2 evaluation. Lower is better for FGD, trajectory/location errors, and average control error. For metrics marked with $\rightarrow$, closer to Ground-Truth is better. \textbf{\textcolor{red}{Red bold}} marks the best generated method in each column, and \underline{\textcolor{blue}{blue underlined}} marks the second best.}
\label{tab:main_results}
\resizebox{\textwidth}{!}{%
\begin{tabular}{lrrrrrr}
\toprule
Method & FGD ($\downarrow$) & BC ($\rightarrow$) & Diversity ($\rightarrow$) & Traj. Error ($\downarrow$) & Loc. Error ($\downarrow$) & Avg. Err. cm ($\downarrow$) \\
\midrule
Ground-Truth & -- & 0.703 & 11.970 & -- & -- & -- \\
\midrule
Inverse Kinematic & 1.043 & \textbf{\textcolor{red}{0.723}} & 16.379 & 0.582 & 0.453 & 19.973 \\
Chunk-Wise ControlNet & 0.491 & 0.778 & \textbf{\textcolor{red}{12.034}} & 0.996 & 0.976 & 46.159 \\
Sequential Chunk Guidance & 0.690 & 0.816 & 13.845 & 0.574 & 0.470 & 11.701 \\
\midrule
\rowcolor{OursGray}
Synchronous Chunk Guidance (Ours) & \underline{\textcolor{blue}{0.442}} & \underline{\textcolor{blue}{0.750}} & 12.234 & \textbf{\textcolor{red}{0.092}} & \textbf{\textcolor{red}{0.100}} & \textbf{\textcolor{red}{4.669}} \\
\rowcolor{OursGray}
\emph{Progressive Chunk Guidance} (Ours) & \textbf{\textcolor{red}{0.431}} & 0.762 & \underline{\textcolor{blue}{11.760}} & \underline{\textcolor{blue}{0.267}} & \underline{\textcolor{blue}{0.161}} & \underline{\textcolor{blue}{6.335}} \\
\bottomrule
\end{tabular}%
}
\end{table*}

Table~\ref{tab:main_results} compares \emph{InteractGesture} across all three chunk-guidance schedules as well as against alternative spatial control paradigms.
Specifically, we compare against a post-hoc \emph{Inverse Kinematics (IK)} baseline, which analytically solves joint rotations post-generation to reach spatial targets without respecting the speech-conditioned gesture distribution, and a \emph{Chunk-Wise ControlNet} baseline, which incorporates spatial conditioning via a trainable feed-forward adapter applied independently to each chunk.
As shown in Table~\ref{tab:main_results}, IK preserves high motion diversity but severely degrades gesture naturalness (FGD $1.043$) and produces large average control errors ($19.97\,\text{cm}$) because geometric post-fitting introduces unnatural body poses.
Chunk-Wise ControlNet achieves better FGD ($0.491$), but fails to reach keyframe targets ($46.16\,\text{cm}$ average error; $97.6\%$ location failure rate).

In contrast, \emph{Sequential Chunk Guidance} improves spatial accuracy ($11.70\,\text{cm}$ error) through test-time latent optimization, though freezing prior chunks still prevents future constraints from adjusting incoming motion trajectories.
\emph{Synchronous Chunk Guidance} achieves the strongest overall accuracy in the offline setting ($4.67\,\text{cm}$ average error; $10.0\%$ location error).
Finally, our proposed \emph{Progressive Chunk Guidance} enables continuous streaming execution while achieving an average error of $6.34\,\text{cm}$ and improving gesture naturalness (FGD $0.431$, Diversity $11.760$) over Synchronous guidance.
Overall, optimizing target latent estimates through inference-time guidance strikes a superior balance between spatial control precision and gesture naturalness compared to post-hoc geometric fitting or feed-forward chunk-level conditioning.


\begin{figure*}[t]
\centering
\includegraphics[width=0.95\textwidth]{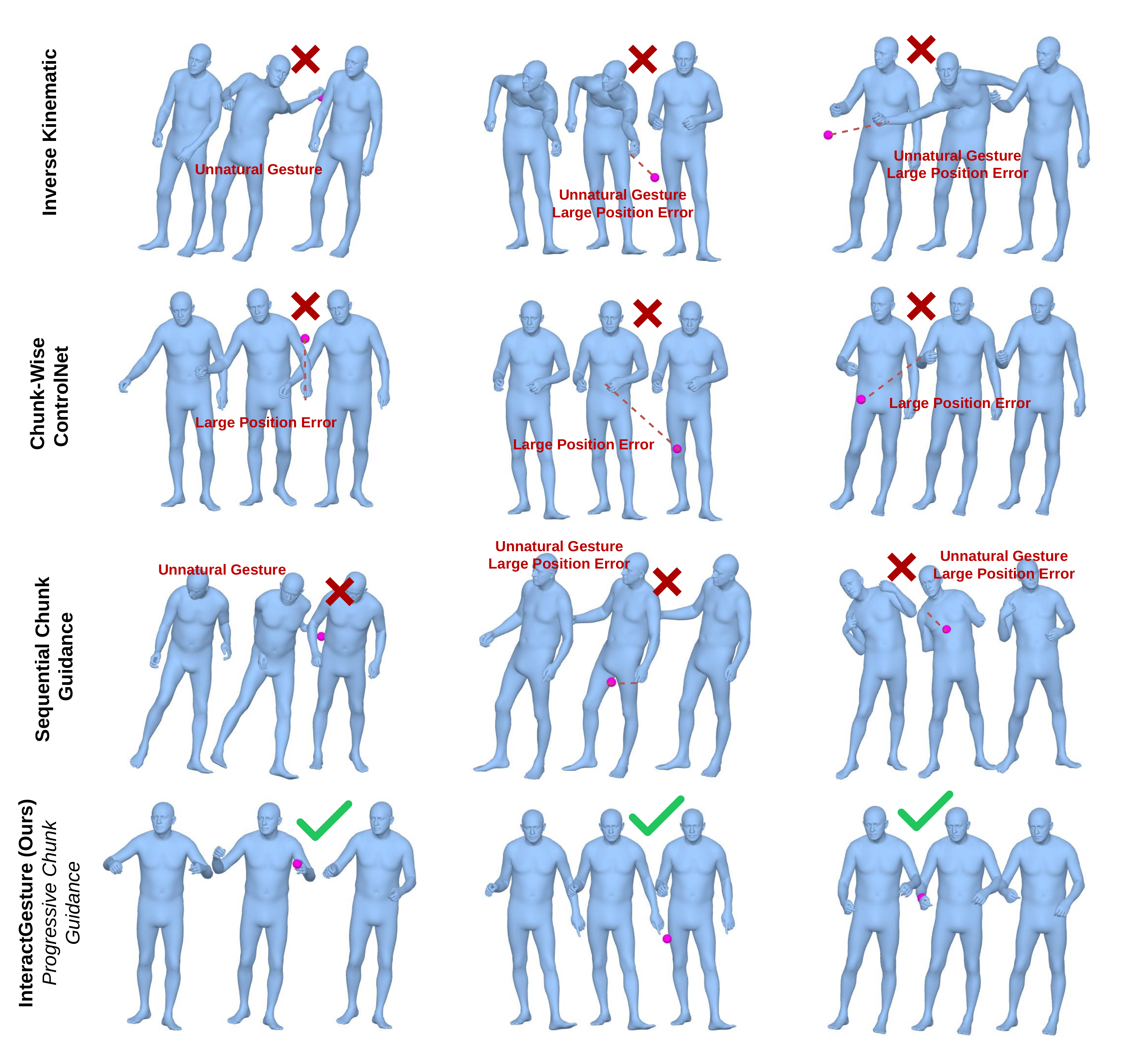}
\caption{
Main qualitative result. \emph{InteractGesture} uses spatial controls to steer co-speech gestures toward target hand and arm positions while preserving the speech-conditioned gesture prior. The visualization shows the guided generations alongside target positions, demonstrating whether the generated wrists and elbows adhere to the requested spatial constraints.
}
\label{fig:main_visualization}
\end{figure*}

\subsection{Qualitative Results}

Figure~\ref{fig:main_visualization} qualitatively compares the visualization of the baselines and our method under identical spatial keyframe controls.
The \emph{Inverse Kinematics} baseline directly calculate joint rotation from control position after generation, without considering the speech-conditioned motion distribution.
As a result, it can place controlled joints closer to the targets, but the surrounding motion often looks less natural because the correction is not regularized by the learned gesture prior.

\emph{Chunk-Wise ControlNet} applies control independently within each generated chunk. Its joint placement is less accurate because a chunk-wise perturbation has limited access to neighboring chunk motion and future control positions, preventing corrections near chunk boundaries from being coordinated over a longer gesture trajectory.
\emph{Sequential Chunk Guidance} improves over chunk-wise control but freezes each previous chunk before optimizing the next one. Therefore, control targets in chunk $k+1$ cannot propagate influence back to chunk $k$, even when the motion should be adjusted across the chunk boundary.
\emph{InteractGesture} with \emph{Progressive Chunk Guidance} keeps overlapping chunks editable as new chunks are introduced. This allows later controls to influence the active previous chunk through refreshed context, producing both accurate spatial control and natural co-speech motion in Figure~\ref{fig:main_visualization}.

\section{Ablation Study}
\label{sec:ablations}

We perform an ablation study on the BEAT2 dataset under absolute 3D spatial control to evaluate the individual contributions of our guidance components, optimization scales, and execution schedules (Table~\ref{tab:ablation_control}).

\begin{table*}[t]
\centering
\caption{Ablation study on spatial guidance components and optimization scales on BEAT2. Panel (a) evaluates guidance modules at scale $0.5$. Panel (b) evaluates guidance scale sweeps. Delay $0$ corresponds to Synchronous Chunk Guidance. Delay $1$ corresponds to \emph{Progressive Chunk Guidance} ($\Delta i = 1$).}
\label{tab:ablation_control}
\resizebox{\textwidth}{!}{%
\begin{tabular}{lrccrrrrrr}
\toprule
Setting & Scale & \begin{tabular}{@{}c@{}}Sampling\\Guidance\end{tabular} & \begin{tabular}{@{}c@{}}Post\\Refine\end{tabular} & FGD ($\downarrow$) & BC ($\rightarrow$) & Div ($\rightarrow$) & Traj10 ($\downarrow$) & Loc10 ($\downarrow$) & Avg. cm ($\downarrow$) \\
\midrule
\rowcolor{OursGray}\multicolumn{10}{c}{\textbf{(a) Control Component Ablation}} \\
\rowcolor{DelayGray}\multicolumn{10}{c}{\emph{Delay 0: Synchronous Chunk Guidance}} \\
Uncontrolled Baseline & 0.5 & $\times$ & $\times$ & 0.439 & \textbf{0.724} & 12.243 & 0.984 & 0.965 & 59.308 \\
Sampling Guidance only & 0.5 & $\checkmark$ & $\times$ & \textbf{0.438} & 0.744 & \textbf{12.175} & 0.426 & 0.427 & 11.471 \\
Post-Refinement only & 0.5 & $\times$ & $\checkmark$ & 0.438 & 0.743 & 12.195 & 0.338 & 0.338 & 9.428 \\
Full Guidance (Ours) & 0.5 & $\checkmark$ & $\checkmark$ & 0.442 & 0.750 & 12.234 & \textbf{0.092} & \textbf{0.100} & \textbf{4.669} \\
\addlinespace[2pt]
\rowcolor{DelayGray}\multicolumn{10}{c}{\emph{Delay 1:} \emph{Progressive Chunk Guidance}} \\
Uncontrolled Baseline & 0.5 & $\times$ & $\times$ & 0.418 & \textbf{0.726} & 12.320 & 0.992 & 0.960 & 64.984 \\
Sampling Guidance only & 0.5 & $\checkmark$ & $\times$ & 0.434 & 0.766 & \textbf{12.151} & 0.733 & 0.484 & 12.740 \\
Post-Refinement only & 0.5 & $\times$ & $\checkmark$ & \textbf{0.417} & 0.746 & 12.272 & 0.524 & 0.524 & 12.361 \\
Full Guidance (Ours) & 0.5 & $\checkmark$ & $\checkmark$ & 0.431 & 0.762 & 11.760 & \textbf{0.267} & \textbf{0.161} & \textbf{6.335} \\
\midrule
\rowcolor{OursGray}\multicolumn{10}{c}{\textbf{(b) Guidance Scale Sweep}} \\
\rowcolor{DelayGray}\multicolumn{10}{c}{\emph{Delay 0: Synchronous Chunk Guidance}} \\
Scale 0.1 & 0.1 & $\checkmark$ & $\checkmark$ & \textbf{0.435} & \textbf{0.739} & 12.224 & 0.299 & 0.376 & 10.464 \\
Scale 0.5 & 0.5 & $\checkmark$ & $\checkmark$ & 0.442 & 0.750 & 12.234 & 0.092 & 0.100 & 4.669 \\
Scale 1.0 & 1.0 & $\checkmark$ & $\checkmark$ & 0.452 & 0.766 & \textbf{11.970} & \textbf{0.080} & \textbf{0.050} & \textbf{4.072} \\
\addlinespace[2pt]
\rowcolor{DelayGray}\multicolumn{10}{c}{\emph{Delay 1:} \emph{Progressive Chunk Guidance}} \\
Scale 0.1 & 0.1 & $\checkmark$ & $\checkmark$ & \textbf{0.417} & \textbf{0.743} & 12.257 & 0.283 & 0.307 & 8.900 \\
Scale 0.5 & 0.5 & $\checkmark$ & $\checkmark$ & 0.431 & 0.762 & \textbf{11.760} & \textbf{0.267} & \textbf{0.161} & \textbf{6.335} \\
Scale 1.0 & 1.0 & $\checkmark$ & $\checkmark$ & 0.463 & 0.778 & 11.283 & 0.355 & 0.191 & 7.308 \\
\bottomrule
\end{tabular}%
}
\end{table*}

\subsection{Impact of Guidance Components}

Table~\ref{tab:ablation_control}(a) evaluates four guidance configurations: Uncontrolled Baseline, Sampling Guidance only, Post-Refinement only, and Full Guidance (Ours). Each configuration is reported for both Delay $0$ and Delay $1$ so that the component effects can be compared under the same execution settings.

Without guidance, spatial control fails in both schedules, with average errors of $59.308\,\text{cm}$ for Delay $0$ and $64.984\,\text{cm}$ for Delay $1$. Sampling Guidance only reduces the error to $11.471\,\text{cm}$ and $12.740\,\text{cm}$, respectively, showing that optimizing target latent estimates during denoising provides the main correction signal. Post-Refinement only also improves accuracy, reaching $9.428\,\text{cm}$ and $12.361\,\text{cm}$, but it lacks the iterative sampler regularization available during denoising.

Full Guidance combines both components and gives the best control accuracy within each delay setting, reducing the average error to $4.669\,\text{cm}$ for Delay $0$ and $6.335\,\text{cm}$ for Delay $1$. This confirms that Sampling Guidance provides coarse trajectory alignment during denoising, while Post-Refinement fine-tunes the final target latent estimates where decoded joint coordinates are most reliable.

\subsection{Guidance Scale Sensitivity}

Table~\ref{tab:ablation_control}(b) evaluates guidance step scales across three settings: $0.1$, $0.5$, and $1.0$.
In the offline Synchronous setting, increasing guidance scale monotonically improves spatial precision, dropping average control error from $10.46\,\text{cm}$ at scale $0.1$ down to $4.07\,\text{cm}$ at scale $1.0$.
For continuous streaming execution under \emph{Progressive Chunk Guidance}, scale $0.5$ provides the optimal trade-off between spatial precision ($6.34\,\text{cm}$) and gesture naturalness (FGD $0.431$), whereas scale $1.0$ yields slightly higher control error ($7.31\,\text{cm}$).

\subsection{Effect of Progressive Chunk Guidance}
\emph{Progressive Chunk Guidance} enables spatial control during continuous streaming generation. Section~\ref{sec:progressive_chunk_guidance} defines the two execution schedules evaluated in Table~\ref{tab:ablation_control}: Delay $0$ corresponds to offline Synchronous Chunk Guidance, which must wait for the full speech sequence so that all chunks can be optimized concurrently; Delay $1$ corresponds to \emph{Progressive Chunk Guidance}, which introduces chunks with a one-wave delay ($\Delta i = 1$) and keeps generation compatible with streaming audio.

The offline Synchronous setting provides the strongest control precision because future controls are available when earlier chunks are still editable. With Full Guidance, it reaches $4.669\,\text{cm}$ average error, Traj10 $0.092$, and Loc10 $0.100$. \emph{Progressive Chunk Guidance} maintains comparable control accuracy under the streaming constraint, with $6.335\,\text{cm}$ average error, Traj10 $0.267$, and Loc10 $0.161$.

\emph{Progressive Chunk Guidance} also maintains high generation quality: its FGD is slightly lower than the Synchronous setting ($0.431$ vs. $0.442$), with Diversity $11.760$ compared with $12.234$. Thus, Synchronous Chunk Guidance serves as an offline accuracy reference, while \emph{Progressive Chunk Guidance} preserves most of the control benefit without requiring the full speech sequence in advance.

\section{Conclusion}

We presented \emph{InteractGesture}, a model-agnostic inference-time method for spatial control of co-speech gesture generation.
The method guides a pretrained co-speech generative sampler by optimizing predicted target latent estimates through a differentiable RVQ-VAE decoder. We proposed \emph{Progressive Chunk Guidance} to support continuous streaming generation: instead of freezing each chunk before the next one is guided, it keeps overlapping chunk latents editable, introduces future chunks with a staggered delay, and refreshes context from the latest guided estimates. This lets spatial constraints in later chunks influence preceding motion across chunk boundaries without requiring the full speech sequence in advance. When the complete audio stream is available, our synchronous setting can produce high-quality motion with precise joint control.
\emph{InteractGesture} supports diverse applications including sparse joint control, dense joint trajectory control, and pointing-direction control.
Experiments on BEAT2 demonstrate that latent guidance significantly reduces joint errors while preserving speech-conditioned gesture naturalness.
\clearpage

\bibliographystyle{splncs04}
\bibliography{main}

\clearpage

\end{document}